\documentclass[11pt,a4paper]{article}
\usepackage[hyperref]{acl2020}
\usepackage{times}
\usepackage{latexsym}

\usepackage{microtype}
\usepackage{times}
\usepackage{helvet}
\usepackage{amsmath}
\usepackage{courier}
\usepackage{graphicx}
\usepackage{mathrsfs}
\usepackage{amsfonts}
\usepackage{CJKutf8}
\usepackage{ulem}
\usepackage{algorithm}% algorithm
\usepackage{algorithmic}% algorithm
\usepackage{float}
\definecolor{Orange}{rgb}{1,0.5,0}

\aclfinalcopy % Uncomment this line for the final submission
\def\aclpaperid{405} %  Enter the acl Paper ID here

\author{
Wangchunshu Zhou$^{1}$\thanks{\ \ This work was done during the first author's internship at Microsoft Research Asia.} ~~~ Tao Ge$^{2}$ ~~~ Chang Mu$^{3}$ ~~~ Ke Xu$^{1}$ ~~~ Furu Wei$^2$ ~~~ Ming Zhou$^2$\\
$^1$Beihang University, Beijing, China\\
$^2$Microsoft Research Asia, Beijing, China\\
$^3$Peking University, Beijing, China\\
{\tt zhouwangchunshu@buaa.edu.cn, kexu@nlsde.buaa.edu.cn}\\
{\tt \{tage, fuwei, mingzhou\}@microsoft.com}\\
{\tt 1801210867@pku.edu.cn}}

\date{}

\begin{document}
\title{
Improving Grammatical Error Correction with \\ Machine Translation Pairs}
%\author{AAAI Press\\
%Association for the Advancement of Artificial Intelligence\\
%2275 East Bayshore Road, Suite 160\\
%Palo Alto, California 94303\\
%}
\maketitle
\begin{abstract}

We propose a novel data synthesis method to generate diverse error-corrected sentence pairs for improving grammatical error correction, which is based on a pair of machine translation models (e.g., Chinese$\to$English) of different qualities (i.e., poor and good). The poor translation model can resemble the ESL (English as a second language) learner and tends to generate translations of low quality in terms of fluency and grammaticality, while the good translation model generally generates fluent and grammatically correct translations. With the pair of translation models, we can generate unlimited numbers of \textit{poor$\to$good} English sentence pairs from text in the source language (e.g., Chinese) of the translators. Our approach can generate various error-corrected patterns and nicely complement the other data synthesis approaches for GEC. Experimental results demonstrate the data generated by our approach can effectively help a GEC model to improve the performance and approaching the state-of-the-art single-model performance in BEA-19 and CoNLL-14 benchmark datasets.

%We build the poor and good translation model with a phrase-based statistical machine translation model with a decreased language model weight and a neural machine translation model respectively. By taking the pair of their translations of the same sentences in a bridge language as error-corrected sentence pairs, we can construct unlimited pseudo parallel data. Our approach is capable of generating diverse fluency-improving patterns without being limited by the pre-defined rule set and the seed error-corrected data. Experimental results demonstrate the effectiveness of our approach and show that it can be combined with other synthetic data sources to yield further improvements.  
%We conduct experiments on both pretraining with synthetic data generated by our approach and the combination with our approach with corruption-based data synthesis method. 
\end{abstract}

\section{Introduction}

Recent work on grammatical error correction (GEC) has proved that synthetic error-corrected data is helpful for improving GEC models~\cite{ge2018fluency,zhao2019improving,lichtarge2019corpora,zhang2019sequence}. However, the error patterns generated by the existing data synthesis approaches tend to be limited by either pre-defined rule sets or the seed error-corrected training data (e.g., for back-translation). To generate more diverse error patterns to further improve GEC training, we propose a novel data synthesis approach for GEC, which employs two machine translation (MT) models of different qualities.

\begin{figure}[t]
    \centering
    \includegraphics[width=1.\linewidth]{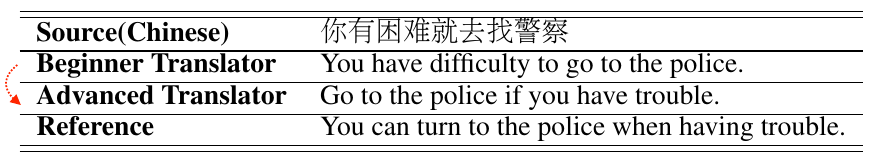}
    \caption{\label{fig:example} Examples of translations generated by the beginner and advanced translator. The beginner translator is implemented with a phrase-based SMT model with the decreased language model weight; while the advanced translator is a state-of-the-art NMT model. The beginner translator tends to literally translate its source language to English, which resembles the way an English learner writes English sentences; while the advanced translator is capable of generating fluent and grammatically correct sentences. By pairing the results of beginner and advanced translators, we can harvest unlimited grammatically improved sentence pairs, as the red dashed arrow shows.}
\end{figure}

%\begin{figure}
%    \centering
%    \includegraphics[width=0.8\linewidth]{gec1.pdf}
%    \caption{Examples of translation results generated by SMT model and NMT model. We can see that SMT %translation resembles that produced by a beginner English learner while NMT translation is grammatically %correct.}
%    \label{fig:m1}
%\end{figure}

%Our method is motivated by the real word English learning scenarios where annotated GEC corpora is collected: An ESL (English as a second language) learner writes sentences which contain grammatical errors and a native English speaker corrects these sentences by producing meaning- preserving and grammatically correct sentences. 
The main idea of our approach is demonstrated in Figure \ref{fig:example}: we use a beginner and an advanced MT model to translate the same sentence in the source language (e.g., Chinese) into English, and pair the poor and good sentence generated by the beginner and advanced translator as an error-corrected sentence pair. This idea is motivated by the studies in English language learning theory~\cite{watcharapunyawong2013thai,bhela1999native,derakhshan2015interference} which find that ESL (English as a second language) learners tend to compose an English sentence by literally translating from their native language with little consideration of the grammar and the expression custom in English.

In our approach, we develop a phrase-based statistical machine translation (SMT) model but decrease its language model weight to make it act as the beginner translator. With the decreased language model weight, the SMT model becomes less aware of the grammar and the expression custom in English, which simulates the behaviors of ESL learners to produce less fluent translations that may contain grammatical errors. On the other hand, we employ the state-of-the-art neural machine translation (NMT) model as the advanced translator which tends to produce fluent and grammatically correct translations. In this way, we can generate diverse error patterns without being limited by the pre-defined rule set and the seed error-corrected data.
We conduct experiments in both the BEA-19~\cite{bryant2019bea} and the CoNLL-14~\cite{ng2014conll} datasets to evaluate our approach. Experiments show the \textit{poor$\to$good} sentence pairs generated by our approach can effectively help a GEC model to improve its performance and achieve the state-of-the-art results in the benchmarks. % With the combination of the proposed data synthesis approach and the existing ones, we achieve the state-of-the-art single model performance on the two datasets, which demonstrates that our approach is able to synthesize high quality pseudo-parallel data for improving the performance of GEC models. We also conduct an in-depth analysis of the improvement yielded by different data synthesis methods and show that our approach can generate more diverse error patterns.
%We evaluate different strategies of using synthetic parallel data generated by the proposed approach, including unsupervised training, fine-tuning, and combination with existing data synthesis methods on GEC.is complementary to existing data synthesis methods and 

Our contributions can be summarized as follows:
\begin{itemize}
	\item We propose a novel data synthesis method to generate diverse error-corrected data for pre-training GEC models based on a pair of machine translation models.
	\item We conduct an empirical study of the commonly used data synthesis approaches for GEC and find their shortcomings in terms of limited error-corrected patterns which can be well addressed by our proposed method.
	\item Our proposed approach can effectively help a GEC model improve its performance and approach the state-of-the-art results in both the CoNLL-14 and the BEA-19 benchmarks.

\end{itemize}

\section{Background: SMT vs NMT}\label{background}

In this section, we briefly introduce both SMT and NMT models and discuss some of their characteristics that motivate the proposed approach.

The phrase-based SMT model is based on the noisy channel model. It formulates the translation probability for translating a foreign sentence $f$ into English $e$ as:
\begin{equation}
    \operatorname{argmax}_{\mathbf{e}} P(\mathbf{e} | \mathbf{f})=\operatorname{argmax}_{\mathbf{e}} P(\mathbf{f} | \mathbf{e}) P(\mathbf{e})
\end{equation}
\noindent where $P(\mathbf{e})$ corresponds to an English language model and $P(\mathbf{f} | \mathbf{e})$ is a separate phrase-based translation model. In practice, an SMT model combines the translation model with a language model with weights tuned through minimum error rate training (MERT)~\cite{och2003minimum} on a validation set. The role of the language model in SMT models is to avoid literal (i.e., phrase-by-phrase) translation and make generated translation more natural and grammatically correct. Without the language model, its produced translation will become less fluent and more likely to contain grammatical errors.

In contrast, a neural machine translation (NMT) model based on sequence-to-sequence architecture is optimized by directly maximizing the likelihood of the target sentences given source sentences $P(\mathbf{e}|\mathbf{f})$. It proves effective to generate adequate and fluent translations, and substantially outperforms SMT models in most cases.

%Similarly, in the GEC domain, recent NMT-based GEC models~\cite{ge2018fluency,zhao2019improving} are able to outperform a carefully designed SMT-based GEC model~\cite{junczys2016phrase}, which held previous state-of-the-art in the GEC task, by a large margin. This also suggests that NMT-based models can generate more fluent outputs that contain less grammatical errors compared with that generated by SMT-based models.

Table \ref{smtnmt} gives a comparison of SMT and NMT in newstest17 Chinese-English news translation dataset. It can be observed that the SMT model is inferior to the NMT model in terms of both the translation quality (reflected by \textbf{BLEU}) and the fluency (reflected by \textbf{Perplexity}\footnote{The perplexity of output sentences is measured by GPT-2~\cite{radford2019language}.}). 

\begin{table}[t!]
	\begin{center}
\scalebox{1.}{
			\begin{tabular}{lcc}
				\hline\hline
				\textbf{Translation Model} & \textbf{BLEU} & \textbf{Perplexity}  \\ \hline
				SMT & 20.3  & 23.1  \\
				%SMT$_{low}$ & 19.4   & 25.5\\
				NMT & 27.2  & 15.7 \\
				\hline\hline
		\end{tabular}}
	\end{center}
	\caption{\label{smtnmt} The performance (i.e. BLEU score) and the fluency of the output sentences (i.e. Perplexity) of the beginner translator (i.e. SMT model) and the advanced translator (i.e. NMT model) used in our experiments on newstest17 Chinese-English translation test set. }
\end{table}

%\begin{equation}
%\text{ER}(S)=\frac{\sum_{i=0}^{n} \text { Distance }\left(S_{i}^{\text {s}}, S_{i}^{\text %{t}}\right)}{\sum_{i=0}^{n} ||S_{i}^{\text {t}}||}
%\end{equation}

%where $n$ is the number of sentence pairs in the corpora S. $S^{\text{s}}$ and $S^{\text{s}}$ is respectively the source and target sentence. While the differences in generated and reference translations may not necessarily be grammatical errors, we expect this metric can still capture the difference in the amount of grammatical errors in translations generated by SMT models and NMT models.

%The comparison results are shown in Table 1. We can see that NMT models outperform SMT models by a large margin, especially when considering the fluency of generated translations. This suggests that translations produced by SMT models may contain much grammar error compared with that generated by NMT models.

%\begin{table}[t!]
%\begin{center}
%\scalebox{1.0}{
%\begin{tabular}{lcc}
%\hline\hline
%\textbf{Metric} & \textbf{SMT} & \textbf{NMT} \\ \hline
%BLEU (Machine Translation) &  & \bf 27.4 \\
%$F_{0.5}$ (Grammar Error Correction) & 50.27  & \bf 56.25 \\
%Perplexity (generated translation) & 5.3  & \bf 2.1 \\
%%Error Rate (generated translation) & 67.8  & \bf 15.3 \\
%\hline\hline
%\end{tabular}}
%\end{center}
%\caption{\label{result1} Comparison between SMT model (Moses~\cite{koehn2007moses}) and NMT model (Transformer). Results on machine translation and grammar error correction are based on reported results on the WMT14 En-Fr test set and CoNLL-2014 shared task on GEC respectively. }
%\end{table}

\begin{figure*}
    \centering
    \includegraphics[width=14cm]{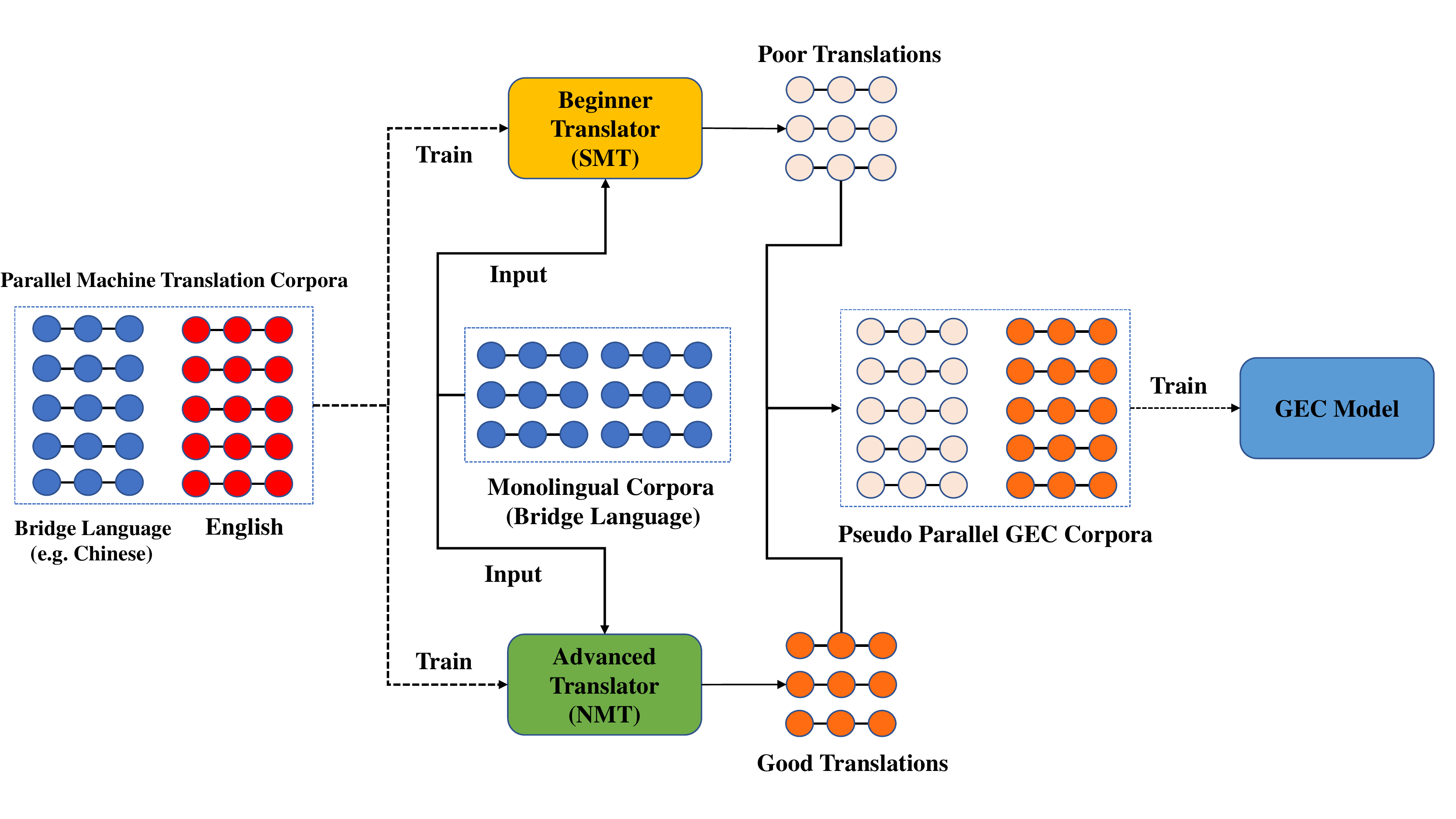}
    \caption{The overview of our approach. Our approach consists of two machine translation models with different qualities that are trained with bi-lingual parallel corpora. The beginner translator and the advanced translator generate poor and good translations respectively. We can thus establish error-corrected data using the \textit{poor$\to$good} translation pairs of sentences in the source language. (Best viewed in color)}
    \label{fig:overview}
\end{figure*}
\section{\textit{Poor$\to$Good} Sentence Pair Generation}

As discussed above, an SMT model is generally inferior to an NMT model in terms of both fluency and translation quality. Motivated by the fact, we propose to employ an SMT model as a \textbf{beginner translator}, and an NMT model as an \textbf{advanced translator}. We use both the translators to translate the same sentences in the source language of the MT models into English, obtaining \textit{poor$\to$good} English sentence pairs. These fluency-improving sentence pairs prove helpful in improving the performance of GEC models, according to the previous work \cite{ge2018fluency,zhang2019sequence}; thus they can be used as augmented data for pre-training a GEC model. The overview of our approach is illustrated in Figure \ref{fig:overview}.

%In this paper, motivated by the data scarcity problem in the GEC task and the comparison between SMT and NMT models, we propose a novel data synthesis method for training GEC models with a pair of \textbf{Beginner Translator} and \textbf{Advanced Translator}. The beginner translator is a relatively poor machine translation model that tends to generate unnatural translations containing many grammatical errors. The advanced translator, in contrast, is well trained and generally outputs fluent and grammatically correct translations. Both translation models are trained to translate sentences from a same bridge language (e.g. Chinese) to the target language in which we want to train the GEC model (e.g. English). After training the pair of machine translation models, we can synthesize pseudo parallel data for training GEC models with the pair of machine translation models. Specifically, we feed sentences from monolingual corpora in the bridge language into the pair of translation models and take the output from the beginner translator and the advanced translator as the source and target sentences for GEC task respectively. We choose Chinese as the bridge language in our experiments as it is less similar to English, thus may be able to cover more error patterns similar to those generated by non-native English speakers. The procedure of the proposed method is illustrated in Figure 2.

\subsection{Poor Sentence Generation}\label{subsec:smt}

To generate poor sentences that contain grammatical errors, we employ a beginner translator, which is implemented through a phrase-based SMT model, to translate sentences from monolingual corpora in the source language (e.g. Chinese) to English. To make the generated sentences less fluent and poor enough, we propose to decrease the tuned language model weight of the SMT model. The resulting beginner translator tends to generate translations that resemble the sentences composed by ESL learners: they translate phrase by phrase from their native (i.e., source) language into English but combine the phrase translations in an unnatural way with little awareness of grammar in English.

We present some samples generated by a automatically tuned SMT model and its counterpart with decreased language model weights respectively in Table \ref{resultdiff}. We can see the translation generated by the SMT model with the decreased language model weight contains more grammatical errors than the automatically tuned SMT model. Such poor sentences can provide more diverse error-corrected learning signals that benefit training a GEC model.

%In addition, based on the observation in previous study~\cite{qiu2019artificial} that synthesized parallel data can better help GEC training when the source sentences are of lower fluency and contain more grammatical errors, we propose to manually reduce the weight of language model in the tuned beginner translator to further reduce the translation quality of the beginner translator. Reducing the language model weight in the beginner translator will result in translations that are less fluent and contain more grammatical errors, thus may help train GEC models better. 

%The SMT model combines a phrase dictionary, which tends to make translation more adequate, and a language model, which makes translation more fluent and grammatically correct, with tuned weights indicating their relative importance.

\begin{CJK*}{UTF8}{gbsn}
\begin{table*}[t!]
\begin{center}
\resizebox{1.\textwidth}{!}{
\begin{tabular}{ll}
\hline\hline
 \textbf{Source(Chinese)} & 两国 复交 符合 两国 人民 的 共同 利益 . \\ \hline
 %\textbf{SMT$_{high}$} & ``In any case , I am satisfied \\
 %& with the performance.'' \\ 
 %\hline
  \textbf{SMT$_{original}$} &  Diplomatic rapprochement between \textbf{the} two countries is the common interest of the peoples of the two countries.  \\ \hline
  \textbf{SMT$_{low}$} & Diplomatic rapprochement between two countries \textbf{in} the common interest of the \textbf{two} peoples of the two countries . \\ \hline
  \textbf{NMT} & The resumption of diplomatic relations between the two countries is in the common interest of the two peoples.\\ \hline \hline
   \textbf{Source(Chinese)} & 以后的生活肯定还会一年比一年好. \\ \hline 
   \textbf{SMT$_{original}$} &  Life will \textbf{be} for \textbf{the} rest of their lives better over \textbf{the} years .  \\ \hline
   \textbf{SMT$_{low}$} &  Life will for the \textbf{their} rest of lives better over years .  \\ \hline
   \textbf{NMT} & Life will definitely be better every year.\\
   
 \hline\hline
\end{tabular}}
\end{center}
\caption{\label{resultdiff} Examples of translation results generated by SMT models with different language model weights and that generated by the NMT model. The differences between the poor sentences generated by SMT$_{original}$ and SMT$_{low}$ are bolded. We can see that translations generated by SMT with decreased language model weight (i.e., SMT$_{low}$) contains more grammatical errors, while the NMT model produces fluent and grammatically correct sentences.}
\end{table*}
\end{CJK*}

\subsection{Good Sentence Generation}

To generate good counterparts for the poor sentences, we use an advanced translator, which is implemented with a state-of-the-art NMT model, to generate good translations from the source sentences. Since both the poor translations and good translations are translated from the same source sentences, they tend to express the same meaning but in different ways. As observed in Table \ref{resultdiff}, compared with the output sentences from the SMT models, the NMT model's outputs are generally more fluent and native-sounding. Therefore, we can pair the poor sentences with the good ones to establish \textit{poor$\to$good} sentence pairs as potentially useful training instances for GEC models.

The aforementioned method can be used to generate poor and good sentences from monolingual corpora in the source language (e.g., Chinese). It can be even more easily used when MT parallel corpora (e.g., Chinese-English) are available. For MT parallel datasets whose source sentences' ground truth English translations are available, we can directly use their ground truth English sentences as the good sentences without the need to use the NMT model to get the good translations from the sources. Since bi-lingual parallel data for MT is much more than that for GEC, it is also a feasible solution to collect the \textit{poor$\to$good} sentence pairs in this way.

%In addition, as parallel corpora for machine translation are generally larger and easier to get than parallel data for GEC models, it would be helpful if this large amount of data can be used for training GEC models. Our method can easily convert parallel corpora between the bridge language and English into GEC training data by using ground-truth translations in the parallel corpora, instead of that generated by the NMT-based advanced translator, as an ``oracle translator''. Synthetic data generated with this approach may contain fewer noise as the targe sentences are human written instead of generated by the NMT model.

%We consider filtered sentences as noisy examples which may confuse the GEC model and reduce the performance. The remaining examples are then considered as true error-corrected examples and used to pretrain GEC models. 

%Following previous work~\cite{qiu2019artificial}, we measure the fluency of a sentence by its perplexity, which is the the inverse probability of the sentence, based on a trained language model. The perplexity is normalized by the number of words in that sentence and a sentence’s and the fluency score is negatively related to its perplexity. 

\section{Models}

\subsection{MT Models}

We train both the beginner translator and the advanced translator with the Chinese-English parallel corpus -- UN Corpus~\cite{ziemski2016united}, which contains approximately 15M Chinese-English parallel sentence pairs with around 400M tokens.

\paragraph{Beginner Translator} We use Moses~\cite{koehn2007moses} to implement the phrase-based SMT model for the beginner translator. Specifically, we use MGIZA++~\cite{gao2008parallel} for word-alignment, and KenLM~\cite{heafield2011kenlm} for training a trigram language model on the target sentences of the Chinese-English parallel data. We tune the weights of each component in the Moses system (e.g. phrase table, language model, etc.) using MERT~\cite{och2003minimum} to optimize the system's BLEU score~\cite{papineni2002bleu} on a development set that is constructed by randomly sampling 5,000 sentence pairs from the parallel corpus. To make the SMT's outputs worse, as Section \ref{subsec:smt} discusses, we decrease the weight of the language model by 20\%. To distinguish the automatically tuned SMT model and the one with the decreased language model weight, we call them SMT$_{original}$ and SMT$_{low}$ respectively.

\paragraph{Advanced Translator Model} We use the Transformer-based NMT model as the advanced translator. Specifically, we use the ``transformer-big'' architecture~\cite{vaswani2017attention}. Chinese sentences are segmented into word-level. Afterward, both Chinese words and English words are split into subwords using the byte-pair encoding technique~\cite{sennrich2015neural}. The vocabulary size is 32K for both Chinese and English. We train the advanced translator with Adam optimizer~\cite{kingma2014adam} with the learning rate 0.0003 and the dropout rate 0.3. We warm-up the learning rate during the first 4K updates and then decrease proportionally to the inverse square root of the number of steps. The resulting model yields a BLEU score of 27.2 on newstest17 Chinese-English translation test set, which is competitive to state-of-the-art results. We use beam search with beam size of 4 when using it to generate good translations.

%which uses 6 layers for both the encoder and the decoder, 16 attention heads, embedding size $d_{model} = 1024$, a position-wise feed-forward network at every layer of inner size $d_{ff} = 4096$
\subsection{GEC Model}

%\todo{clearly state that you use the synthesized data for pre-training and fine-tuning on the gold training set. Otherwise, no one knows how you used the synthesized data.}

As for the GEC model, we use the same ``transformer-big'' architecture as our GEC model with tied output layer, decoder embedding, and encoder embeddings. Both input and output sentences are tokenized with byte-pair encoding with shared codes consisting of 30,000 token types. Following the previous work~\cite{zhang2019sequence}, we train the GEC models on 8 Nvidia V100 GPUs, using Adam optimizer~\cite{kingma2014adam} with $\beta_{1}$=0.9, $\beta_{2}$=0.98. We allow each batch to have at most 5,120 tokens per GPU. During pre-training, we set the learning rate to 0.0005 with linear warm-up for the first 8k updates, and then decrease proportionally to the inverse square root of the number of steps. For fine-tuning, the learning rate is set to 0.0001 with warmup over the first 4,000 steps and inverse square root decay after warmup. The dropout ratio is set to 0.2 in both pre-training and fine-tuning stages. We pre-train the model for 200k steps and fine-tune it up to 50k steps.

We use the synthesized data generated with different data synthesis approaches for pre-training GEC models. Then, we use the GEC training data (see Section \ref{subsec:data}) to fine-tune the pre-trained models. We select the best model checkpoint according to the perplexity on the BEA-19 validation set for both the pre-training and fine-tuning. We use beam search to decode with the beam size of 12.

\section{Experiments}

\subsection{Data}\label{subsec:data}

Following the previous work~\cite{grundkiewicz2019neural,choe2019neural,kiyono2019empirical} in GEC, the GEC training data we use is the public Lang-8~\cite{mizumoto2011mining}, NUCLE~\cite{dahlmeier2013building}, FCE~\cite{yannakoudakis2011new} and W\&I+LOCNESS datasets~\cite{bryant2019bea,granger1998computer}.

To generate the \textit{poor$\to$good} English sentence pairs that benefit GEC training, we collect monolingual Chinese news corpora -- Chinese Gigaword~\cite{graff2005chinese} and news2016zh~\cite{chinese} -- to generate poor and good English translations using the SMT and NMT model respectively. After filtering\footnote{We discard the sentence pairs whose edit rate (i.e., the edit distance normalized by the source sentence's length) is larger than 0.6.}, we obtain 15M \textit{poor$\to$good} English sentence pairs. In addition, we generate 15M poor English sentences from the Chinese-English parallel corpus -- UN Corpus -- by translating the Chinese sentences with the beginner translator, and then pair them with their ground truth English translations as \textit{poor$\to$good} sentence pairs. %Specifically, we synthesize around 50M sentence pairs with the pair of the beginner translator (with decreased language model score) and the advanced translator as pseudo-parallel data for GEC training. With available Chinese-English parallel corpora, we additionally synthesize 25M sentence pairs by pairing the beginner translator and ground-truth translation as pseudo-parallel sentence pairs.

%Empirically, we find that the output translations generated by the SMT model sometimes contain too many errors, which may introduce too many noises in the synthesized data and affect the GEC model's performance.  After filtering, we use the remaining 15M sentence pairs generated by the SMT-NMT pair and 15M sentence pairs from SMT and the ground truth translations for pretraining GEC models.

We additionally include 60M sentence pairs obtained by the corruption based approach ~\cite{zhao2019improving}, 60M pairs from the round-trip translation approach~\cite{lichtarge2019corpora}, and 60M pairs by the fluency boost back-translation approach \cite{ge2018fluency} for GEC pre-training. Specifically, the corruption-based and round-trip translation data is obtained from the NewsCrawl dataset; while the back-translated data is harvested from English Wikipedia with a backward model trained on the public Lang-8 and NUCLE dataset.
\begin{table}[t!]
\begin{center}
\resizebox{1.\linewidth}{!}{
\begin{tabular}{lcc}
\hline\hline
\textbf{Dataset} & \textbf{\#sent(pairs)} & \textbf{Split} \\ \hline
\bf SMT-NMT pairs & 15M &  pre-train \\
\bf SMT-gold pairs & 15M &  pre-train \\
\bf corruption & 60M &  pre-train \\
\bf back-translation & 60M &  pre-train \\ \hline
\bf Lang-8 & 1.04M &  fine-tune \\ 
\bf NUCLE & 57.1K &  fine-tune \\
\bf FCE & 28.4K &  fine-tune \\
\bf W\&I train & 34.3K & fine-tune \\ \hline
\bf W\&I valid & 4,384 & valid \\
\bf W\&I test (BEA-19) & 4,477 & test \\
\bf CoNLL-14 & 1,312 & test \\
\hline\hline
\end{tabular}}
\end{center}
\caption{\label{data} Statistics of the datasets used for pre-training and fine-tuning.}
\end{table}

We evaluate the performance of GEC models on the BEA-19 and the CoNLL-14 benchmark datasets. 
Following the latest work in GEC~\cite{lichtarge2019corpora,zhao2019improving,grundkiewicz2019neural,kiyono2019empirical,choe2019neural,zhou2020pseudo,omelianchuk2020gector}, we evaluate the performance of trained GEC models using $F_{0.5}$ on test sets using official scripts\footnote{M2scorer for CoNLL-14; Errant for BEA-19.} in both datasets. The data sources used for pretraining, fine-tuning, and evaluating the GEC models are summarized in Table \ref{data}.

%As our primary goal is to explore and analyze the effect of synthetic parallel data generated by the proposed approach, we do not incorporate extensive tricks including iterative decoding, edit-weighted MLE objective, right-to-left re-ranking, external spell checker, etc. We compare the proposed data synthesis method against the most commonly used data synthesis approach for GEC, including the corruption-based approach and the back-translation method. 

%\begin{itemize}
    %\item unsupervised training of GEC model solely with data generated with our approach, denoted by %GEC$_{ours}$
    %\item unsupervised training with the combination of randomly corrupted monolingual data and data %generated with our approach, denoted by GEC$_{ours+corr}$
    %\item supervised training exclusively with parallel GEC corpora, denoted by GEC$_{train}$
    %\item fine-tuning GEC model with parallel GEC data based on GEC model pretrained solely with our %approach, denoted by GEC$_{ours+fine}$.
    %\item fine-tuning GEC model with parallel GEC data based on GEC model pretrained with the combination %of randomly corrupted monolingual data and data generated with our approach, denoted by GEC$_{all}$.
%\end{itemize}
%\subsection{Experimental Results}

\subsection{Results}
Table \ref{result3} compares the performance of our model to the previous studies in the same test sets. It is notable that all the results in Table \ref{result3} are the single model's result with beam search decoding. We do not compare to the results obtained with additional inference methods like iterative decoding and reranking with an LM or a right-to-left GEC model, because they are not related to our contributions.

\begin{table}[!t]
\begin{center}
\resizebox{1.\linewidth}{!}{
\begin{tabular}{lcc}
\hline\hline
\textbf{Method} & \textbf{BEA-19}& \textbf{CoNLL-14} \\ \hline
\bf \citet{junczys2018approaching} & - & 53.0 \\
\bf \citet{lichtarge2019corpora} & - & 56.8 \\
\bf \citet{zhao2019improving}  & - & 59.8 \\
\bf \citet{choe2019neural}  & 63.1 & 60.3 \\
\bf \citet{kiyono2019empirical} & 64.2 & 61.3 \\ \hline
\bf Ours & \bf 65.2 & \bf 62.1 \\
%\bf \citet{junczys2018approaching} & ensemble \times 4 & 55.8 \\
%\bf \citet{lichtarge2019corpora} & ensemble $\times$ 8 + iterative decoding & 60.4 \\

%\bf \citet{grundkiewicz2019neural} & ensemble $\times$ 4 + iterative decoding + LM + R2L  & 64.2 \\
%\bf \citet{kiyono2019empirical} & ensemble $\times$ 4 + SSE +R2L  & \bf 65.0 \\

%\bf Ours  & ensemble x 4  &  \\

\hline\hline
\end{tabular}}
\end{center}
\caption{\label{result3} The comparison of our single model against the state-of-the-art single models in the previous work in the BEA-19 and CoNLL-2014 test set. ``-'' denotes that the previous work does not report its single model's performance in the test set. It is notable that among the previous work in this table, the first three do not use W\&I+LOCNESS for training the model. We do not compare with the work that does not report its single model's performance in either of the test sets.}
\end{table}

According to Table \ref{result3}, our model outperforms the state-of-the-art single model results in both BEA-19 and CoNLL-14 test set. The main difference between our model and the previous state-of-the-art models is that we additionally use the \textit{poor$\to$good} English sentence pairs obtained from the pair of MT models, accounting for the improvement over the previous work.

\begin{table}[t!]
\begin{center}
\resizebox{1.\linewidth}{!}{
\begin{tabular}{lcc}
\hline\hline
\textbf{Method} & \textbf{BEA-19} & \textbf{CoNLL-14} \\ \hline
\bf Baseline & 57.1 & 51.5 \\ \hline
\multicolumn{3}{c}{Pre-train with 30M synthesized data \& fine-tune} \\ \hline
\bf Corr(30M) & 59.5 &  55.7 \\
\bf RT(30M) & 58.9 &  55.2 \\ 
\bf BT(30M) & 59.4 &  55.9 \\
\bf Ours(30M) & \bf 60.4 & \bf 56.6 \\\hline
\multicolumn{3}{c}{Pre-train with 60M synthesized data  \& fine-tune} \\ \hline
\bf Corr(60M) & 59.9 & 55.9 \\ 
\bf RT(60M) & 59.7 & 55.8 \\ 
\bf BT(60M) & 60.5 & 56.5 \\
\bf Corr(30M) + RT(30M) & 61.2 & 57.1 \\ 
\bf Ours(30M)+Corr(30M)  & 61.9 & 57.7 \\
\bf Ours(30M)+BT(30M)  &  \bf 63.1 & \bf 58.5 \\
\hline\hline
\end{tabular}}
\end{center}

\caption{\label{resultfine} The performance of GEC models pre-trained with various synthesized data and fine-tuned with the GEC training data. \textbf{Ours} denotes the synthesized data generated with our approach, \textbf{Corr},\textbf{RT}, and \textbf{BT} denotes the synthesized data generated by random corruption, round-trip translation, and back-translation. \textbf{Baseline} denotes the GEC model directly trained with the GEC training data without pre-training.}
\end{table}

\begin{table}[t!]
\begin{center}
\resizebox{1.\linewidth}{!}{
\begin{tabular}{lcc}
\hline\hline
\textbf{Method} & \textbf{BEA-19} & \textbf{CoNLL-14} \\ \hline
\multicolumn{3}{c}{Pre-train with 30M synthesized data} \\ \hline
\bf Corr(30M) & 37.1 &  27.5 \\
\bf Round-trip(30M) & 39.7 &  29.2 \\
\bf BT(30M) & \bf 45.1 &  \bf 33.2 \\
\bf Ours(30M) & 43.5 & 31.1  \\ \hline
\multicolumn{3}{c}{Pre-train with 60M synthesized data} \\ \hline
\bf Corr(60M) & 38.7 & 28.1 \\ 
\bf RT(60M) & 40.2 & 29.8 \\
\bf BT(60M) & 47.4 & 35.5 \\ 
\bf Corr(30M) + RT(30M) & 47.2 & 34.7 \\ 
\bf Ours(30M)+Corr(30M)  &  45.6  &  33.9 \\
\bf Ours(30M)+BT(30M)  & \bf 49.5  & \bf 36.2 \\
\hline\hline
\end{tabular}}
\end{center}
\caption{\label{resultunsupervised} The performance of GEC models pretrained with different synthesized data without fine-tuning on GEC training data.}
\end{table}

To conduct an in-depth analysis of the improvement by the synthesized data, we compare the performance of GEC models pre-trained with different data sources. According to Table \ref{resultfine}, the model pre-trained with the 30M synthesized sentence pairs from the beginner and advanced translator outperforms its counterparts that are pre-trained with the same amount of data synthesized from back-translation, round-trip translation, and corruption-based approaches, demonstrating that our approach provides more valuable and diverse error-corrected learning signals for GEC models. When we increase the amount of the synthesized data for pre-training to 60M, we observe that the models pre-trained with a single data source (i.e., \textbf{Corr}, \textbf{BT},and \textbf{RT}) improve only a little over their 30M counterparts, indicating that their generated error-corrected patterns are limited, which is consistent with the observation of the previous studies \cite{edunov2018understanding}. In contrast, if combining multiple data sources for pre-training, the performance will be significantly improved. Among them, the GEC models pre-trained with the \textit{poor$\to$good} sentence pairs yield the best results, which demonstrates that the error-corrected patterns provided by the \textit{poor$\to$good} sentence pairs generated through our approach are different from those by back-translation and corruption-based approaches and they can nicely complement each other to achieve a better result.

%We then compare the GEC models trained with different combination of synthetic data sources. First, we find that the combination of the back-translation based approach with other data synthesis methods substantially improves the performance, out-performing that simply double the data size from one synthetic data source (i.e. \textbf{BT(60M)}). This demonstrates that the combination of different synthetic data sources may introduce different error patterns that are complementary with each other and improve the coverage of error types, thus is more effective than simply increasing the data size from one synthetic source. We also find the combination of synthetic data generated by our approach with the other two sources can yield a larger improvement, which confirms that our approach can generate more diverse error-patterns than existing data synthesis approaches. 

%In addition, the influence of the language model weight in the beginner translator on the final performance of GEC models are consistent with that found in the unsupervised training results, which confirms that decreasing automatically tuned language model weight in the beginner translator may help train GEC models better.  We think that it is because synthetic data generated by back-translation method contains similar error-patterns to that in the parallel GEC data, which makes it less effective when parallel data is available. 

\begin{table}[!t]
	\begin{center}
\resizebox{1.\linewidth}{!}{
			\begin{tabular}{lcc}
				\hline\hline
				\textbf{Method} & \textbf{BEA-19} & \textbf{CoNLL-14} \\ \hline
				\multicolumn{3}{c}{\textbf{Pre-training Only}} \\ \hline
				\bf Ours(30M) & \bf 43.5 & \bf 31.1  \\
				~ - w/o SMT-NMT (15M) & 40.1 & 28.8 \\
				~ - w/o SMT-gold (15M) & 39.5 & 28.4 \\ \hline
				%~ - w/o fluency filtering & 35.4 & 25.7 \\ \hline
\multicolumn{3}{c}{\textbf{Pre-training + Fine-tuning}} \\ \hline
				\bf Ours(30M) & \bf 60.4 & \bf 56.6 \\
				~ - w/o SMT-NMT (15M) & 58.2 & 55.2 \\
				~ - w/o SMT-gold (15M) & 57.8 & 54.8 \\
				%~ - w/o fluency filtering & 56.6 & 49.1\\
				\hline\hline
		\end{tabular}}
	\end{center}
	\caption{\label{resultablated} The ablation study for comparing the contribution of the SMT-NMT and SMT-gold pairs to the final results.}
\end{table}

\begin{table}[!t]
	\begin{center}
\resizebox{1\linewidth}{!}{
			\begin{tabular}{lcc}
				\hline\hline
				\textbf{Method} & \textbf{BEA-19} & \textbf{CoNLL-14} \\ \hline
				\multicolumn{3}{c}{\textbf{Pre-training Only}} \\ \hline
				\bf Ours(30M) & \bf 43.5 & \bf 31.1  \\
				~ - w/o decreased LM score & 42.7 & 30.3 \\ \hline
				%~ - w/o fluency filtering & 35.4 & 25.7 \\ \hline
\multicolumn{3}{c}{\textbf{Pre-training + Fine-tuning}} \\ \hline
				\bf Ours(30M) & \bf 60.4 & \bf 56.6 \\
				~ - w/o decreased LM score & 59.1 & 55.8 \\
				%~ - w/o fluency filtering & 56.6 & 49.1\\
				\hline\hline
		\end{tabular}}
	\end{center}
	\caption{\label{tab:lowlm} The ablation study to test the effect of decreasing the language model weight of the SMT model in the final results.}
\end{table}

Moreover, we test the performance of the pre-trained models without fine-tuning to see the quality of the synthesized data. According to Table \ref{resultunsupervised}, among the 30M single data sources, the data generated by back translation yields the best performance in both test sets, because back translation introduces informative error-corrected patterns with much less undesirable noise than the corruption-based approach and our approach. When we double the data size for pre-training, we observe the similar results to those in Table \ref{resultfine}: the combinations of different sources of the synthesized data lead to the best results, verifying our assumption that the error-corrected patterns of different data sources are different and they are complementary.

\begin{CJK*}{UTF8}{gbsn}
	\begin{table*}[!ht]
		\centering
\resizebox{1.\textwidth}{!}{
				\begin{tabular}{ll}
					\hline\hline
					\textbf{Source Sentence} & ``我们应该保持身体健康'' \\ \hline
					\textbf{Ground-truth Translation} & ``We should stay healthy.'' \\ \hline
					\textbf{Translation from NMT} & ``We should stay healthy.'' \\ \hline
					%\textbf{Translation from SMT$_{high}$} & ``In order to participate, in advance to the list .'' \\ \hline
					%\textbf{Translation from SMT$_{original}$} & ``For ensure that your regular participants, please sign up the name in advance '' \\ \hline
					\textbf{Translation from SMT} & ``We should \uline{keep} \uline{a} \uline{body} healthy.'' \\ \hline
					\textbf{Rule-based Corruption} & ``We \rule[-2pt]{0.3cm}{0.5pt} \uline{stays} \uline{stay} healthy.'' \\ \hline
					\textbf{Back Translation} & ``We should \rule[-2pt]{0.3cm}{0.5pt} healthy.'' \\ \hline
					\textbf{Round-trip Translation} & ``We should stay healthy.'' \\ \hline\hline
					\textbf{Source Sentence} & ``无论 如 何 , 我 对 大家 的 表现 都 很 满意'' \\ \hline
					\textbf{Ground-truth Translation} & ``Anyway , I am very satisfied with everyone's performance .'' \\ \hline
					\textbf{Translation from NMT} & ``Anyway , I am satisfied with everyone's performance .'' \\ \hline
					%\textbf{Translation from SMT$_{high}$} & ``In any case , I am satisfied with the performance.'' \\ \hline
					%\textbf{Translation from SMT$_{original}$} & ``In any event , I have everyone is very satisfactory performance .'' \\ \hline
					\textbf{Translation from SMT} & ``Regardless of whether such to what , I am very satisfied with both the performance of together.'' \\ \hline
					\textbf{Rule-based Corruption} & ``Anyway , I \rule[-2pt]{0.3cm}{0.5pt} very \uline{satisfy} with \uline{with} everyone's \rule[-2pt]{0.3cm}{0.5pt} .'' \\ \hline
					\textbf{Back Translation} & ``Anyway , I \uline{was} satisfied with \uline{everyone} performance .'' \\ \hline
					\textbf{Round-trip Translation} & ``Anyway , I am very satisfied with everyone's performance .'' \\ \hline\hline
					\textbf{Source Sentence} & ``在 大众 并 不 了解 AI 技术 时 更 是 如此 .'' \\ \hline
                    \textbf{Ground-truth Translation} & ``This is especially true when the public does not understand AI technology .'' \\ \hline
					\textbf{Translation from NMT} & ``This is particularly true when the public does not understand AI technology .'' \\ \hline
					%\textbf{Translation from SMT$_{high}$} & ``In general it is truth and unaware of AI.'' \\ \hline
					%\textbf{Translation from SMT$_{original}$} & ``It is this public when the technology and knowledge of AI is not understand. '' \\ \hline
					\textbf{Translation from SMT} & ``Popular understanding of AI technology not at the time is even more true .'' \\ \hline
					\textbf{Rule-based Corruption} & ``This \uline{are} especially \rule[-2pt]{0.3cm}{0.5pt} when \rule[-2pt]{0.3cm}{0.5pt} public \uline{not} \uline{do} understand AI \uline{AI} technology .'' \\ \hline
                    \textbf{Back Translation} & ``This is especially true when \rule[-2pt]{0.3cm}{0.5pt} public \uline{do} not understand \uline{the} AI technology .'' \\ \hline
                    \textbf{Round-trip Translation} & ``This is \uline{particularly} true when the public does not understand AI technology .'' \\
                    \hline\hline
			\end{tabular}}
		\caption{\label{qualitative} Examples of translations generated by the beginner translator and the advanced translator, together with synthetic erroneous sentences generated by existing approaches. }
	\end{table*}
\end{CJK*}

\subsection{Analysis of the synthesized \textit{poor$\to$good} sentence pairs}

As mentioned in Section \ref{subsec:data}, among the 30M synthesized \textit{poor$\to$good} sentence pairs, 15M are SMT$\to$NMT pairs, while the others are SMT$\to$ground-truth translation pairs. We perform an ablation study to analyze how much they separately contribute to GEC models. According to Table \ref{resultablated}, the sentence pairs with ground truth translation as their good sentences yield better results. The reason is easy to understand: the quality of the ground truth translations is generally better than that of the advanced translator. However, given the fact that bi-lingual parallel data is much less than the monolingual text data, it is more practical to use the beginner and advanced translator to generate the \textit{poor$\to$good} sentence pairs that will not be limited by bi-lingual parallel corpora. Moreover, the results in Table \ref{tab:lowlm} show that decreasing the language model weight leads to more than 1.0 absolute improvement in $F_{0.5}$ score in both test sets, because it can help the SMT model to act more like a beginner translator, which can also be illustrated by the examples in Table \ref{resultdiff}.

We also conduct a qualitative analysis of different data synthesis approaches in Table \ref{qualitative}. We can see that the beginner translator (i.e., SMT) generates less fluent English sentences by literally (phrase by phrase) translating the source sentence, which resembles how the ESL learners write an English sentence. The advanced translator (i.e., NMT) generates high-quality English sentences that can be comparable to the ground-truth translations. Such diverse \textit{poor$\to$good} sentence pairs are helpful to teacher a GEC model how to rewrite a poor sentence into a good one, accounting for the improved results we achieved. In contrast, existing data synthesis approaches such as rule-based corruption and back-translation tends to generate similar error patterns such as verb/noun forms and word deletions, while the round-trip translation method generates limited modifications which are often paraphrase-like and grammatically correct.

\begin{table}[!t]
	\begin{center}
\resizebox{1\linewidth}{!}{
			\begin{tabular}{lcc}
				\hline\hline
				\textbf{Method} & \textbf{Error Rate} & \textbf{\% Error in Rules} \\ \hline
				\bf Real Data &  21.3 & 87.2  \\				
				\bf Ours &  45.3 & 68.6  \\
				\bf Rule-based Corruption &  40.3 & 100 \\
				\bf Round-trip Translation &  6.2 & 61.7 \\
				\bf Back-translation & 25.7  & 99.2 \\
				\hline\hline
		\end{tabular}}
	\end{center}
	\caption{\label{tab:error-analysis} Error-type analysis of different data sources. Error in rules represent the ratio of errors that are \textbf{not} noted as "other" or "unknown" errors by ERRANT.}
\end{table}

We then conduct an in-depth analysis of the error-type contrained in the synthetic data generated by our approach and other data synthesis methods using ERRANT~\cite{bryant-etal-2017-automatic}. The result is in \ref{tab:error-analysis}. We find that almost all error generated by rule-based corruption and back-translation are included in the pre-defined rules in ERRANT system, while many errors in the real datasets are beyond these rules. This mismatch in the distribution of error types can often severely impact the performance of data synthesis techniques for grammar correction~\cite{felice2014grammatical}. In contrast, our method can generate much more diverse error patterns that are not limited by the pre-defined error types, which may account for the performance gain.

\section{Related Work}

Grammatical error correction (GEC) is a well-established natural language processing (NLP) task that aims to build systems for automatically correcting errors in written text, particularly in non-native written text. While recently sequence tagging~\cite{ribeiro2018local,omelianchuk2020gector} or word substitution with pre-trained language model~\cite{zhou2019bert,li2020towards} based GEC models have  shown promising results and improved efficiency, seq2seq-based GEC models still remain to be the mainstream method for automated gramamtical error correction. However, as shown in Table \ref{data}, the combination of parallel GEC corpora only yields less than 1.5M sentence pairs, which makes it hard to train large neural models (e.g. transformers) to achieve better results. Prior studies~\cite{rei2017artificial,xie-etal-2018-noising,ge2018fluency,zhao2019improving,kiyono2019empirical} have investigated various approaches to alleviate the data scarcity problem for training GEC models by synthesizing pseudo-parallel GEC data for pretraining GEC models.
We introduce the most commonly used data synthesis approaches for pretraining GEC models and discuss their pros and cons in this section.

\paragraph{Rule-based Monolingual Corpora Corruption} A straightforward data synthesis method is to corrupt monolingual corpora with either pre-defined rules or errors extracted from the seed parallel GEC data~\cite{foster2009generrate,zhao2019improving,wang2019controllable}. The advantage of this approach is that it is very simple and efficient for generating pseudo-parallel GEC data from monolingual corpora. However, manually designed rules are limited and only cover a small portion of grammatical error types written by ESL learners. This makes the improvement yielded by pretraining exclusively with synthetic data generated by this approach very limited, which is demonstrated in our experiments.

\paragraph{Back-translation based Error Generation} This approach trains an error generation model by using the existing error-corrected corpora in the opposite direction and introduces noise into a clean corpus~\cite{rei2017artificial,xie-etal-2018-noising,ge2018fluency}. Concretely, an error generation model is trained to take a correct sentence as input and outputs an erroneous version of the original sentence. It is used to synthesize error-corrected data by taking monolingual corpora as input. This approach is able to cover more diverse error types compared with rule-based corruption method. However, it requires a large amount of annotated error-corrected data, which is not always available, to train the error generation model. In addition, the error patterns generated by this method are generally limited to that contained in the GEC parallel data, which makes the effect of synthetic data generated by back-translation quickly saturates as the amount of synthetic data grows, as demonstrated in our experiments.

%hinders the application of this approach for generating pseudo-parallel GEC data unsupervisedly in specific domains or other Languages

\paragraph{Data Generation from Wikipedia Revision} This approach is based on revision histories from Wikipedia~\cite{cahill2013robust,lichtarge2019corpora}. Specifically, it extracts source-target pairs from Wikipedia edit histories by taking two consecutive snapshots as a single revision to the page to form the error-corrected sentence pairs. This method is able to collect human-made revisions that may better resemble real error-corrected data. However, the vast majority of extracted revisions are not grammatical error corrections, which makes the synthesized data noisy and requires sophisticated filtering before used for pre-training. In addition, the domain of available revision history is limited, which makes this method less general compared to other approaches that can generate synthetic data using any monolingual corpora.
%~\citet{cahill2013robust} show that preposition corrections extracted from Wikipedia revisions improve the quality of a GEC model for correcting preposition errors. Afterward,~\citet{lichtarge2019corpora} 
\paragraph{Data Generation from Round-trip Translations} Round-trip translation~\cite{desilets2009using,madnani-etal-2012-exploring,lichtarge2019corpora} is an alternative approach to synthesis pseudo-parallel data for GEC pre-training with monolingual corpora. This approach attempts to introduce noise via bridge translations. Specifically, it uses two state-of-the-art NMT models, one from English to a bridge language and the other from the bridge language to English. With the MT models, it takes the original sentences from monolingual corpora as the target sentences, and takes the outputs of the round-trip translation as the corresponding source sentences. However, when good translation models are employed, as in the case of~\cite{lichtarge2019corpora}, the resulting source sentences are very likely to be clean and without grammatical errors; on the other hand, when poor machine translation models, such as the SMT models with decreased language model weight, are employed, it may result in a semantic drift from target sentences because two consecutive low quality translation are made, which is undesirable for training GEC models. 
%In addition, we observe that the noises introduced by round-trip translation are often information loss rather than grammatical errors.  
In contrast, the difference between the two machine translation models employed in our approach ensures that the source sentences are of low fluency and contain many grammatical errors.

\section{Conclusion and Future Work}

We propose a novel method to synthesize \textit{poor$\to$good} sentence pairs for pre-training GEC models based on a pair of MT models of different quality. The generated sentence pairs contain diverse error-corrected patterns that can nicely complement other data augmentation approaches, leading to a performance approaching the state-of-the-art single model results in GEC benchmarks. For future work, we plan to investigate the influence of different source languages of the MT models in the performance of GEC, which might be helpful in building a customized English GEC model for the people speaking a specific foreign language.

\section*{Acknowledgments}
We thank the anonymous reviewers for their valuable comments.
\bibliography{acl2020}
\bibliographystyle{acl_natbib}
\end{document}